%% file: main.tex
\documentclass[10pt,twocolumn,letterpaper]{article}

\usepackage{cvpr}              
\usepackage{booktabs}
\input{preamble}

\definecolor{cvprblue}{rgb}{0.21,0.49,0.74}
\usepackage[pagebackref,breaklinks,colorlinks,allcolors=cvprblue]{hyperref}

\def\paperID{4669} 
\def\confName{CVPR}
\def\confYear{2025}

\newcommand\blfootnote[1]{
    \begingroup
    \renewcommand\thefootnote{}\footnote{#1}
    \addtocounter{footnote}{-1}
    \endgroup
}

\title{\modelname: Enhanced Diffusion for High-quality Large-motion \\ Video Frame Interpolation}

\author{Zihao Zhang$^{1,2}$\quad Haoran Chen$^{1,2}$\quad Haoyu Zhao$^{1,2}$\quad Guansong Lu$^3$\\
Yanwei Fu$^{1,2}$\quad Hang Xu$^3$\quad
Zuxuan Wu$^{1,2, \textrm{\Letter}}$\\
$^{1}$Shanghai Key Lab of Intell. Info. Processing, School of CS, Fudan University
    \\
$^{2}$Shanghai Collaborative Innovation Center of Intelligent Visual Computing
\\
$^{3}$Noah's Ark Lab, Huawei
\\
\url{https://github.com/bbldCVer/EDEN}
}

\begin{document}
\maketitle
\input{sec/0_abstract}    
\input{sec/1_intro}
\input{sec/2_related_work}
\input{sec/3_method}
\input{sec/4_experiments}

\input{sec/5_conclusion}

{
    \small
    \bibliographystyle{ieeenat_fullname}
    \bibliography{main}
}

\input{sec/X_suppl}

\end{document}

%% file: preamble.tex
\newcommand{\eps}{{\boldsymbol \varepsilon}}
\newcommand{\modelname}{EDEN}
\usepackage{multirow}
\usepackage{graphicx}
\usepackage{booktabs}
\usepackage{tikz}
\usepackage[normalem]{ulem}
\useunder{\uline}{\ul}{}
\usepackage[misc]{ifsym}
\usepackage[hang]{footmisc}

%% file: sec/0_abstract.tex
\begin{abstract}

Handling complex or nonlinear motion patterns has long posed challenges for video frame interpolation. Although recent advances in diffusion-based methods offer improvements over traditional optical flow-based approaches, they still struggle to generate sharp, temporally consistent frames in scenarios with large motion. To address this limitation, we introduce \textbf{\modelname}, an \textbf{E}nhanced \textbf{D}iffusion for high-quality large-motion vid\textbf{E}o frame i\textbf{N}terpolation. Our approach first utilizes a transformer-based tokenizer to produce refined latent representations of the intermediate frames for diffusion models. We then enhance the diffusion transformer with temporal attention across the process and incorporate a start-end frame difference embedding to guide the generation of dynamic motion. Extensive experiments demonstrate that \modelname~achieves state-of-the-art results across popular benchmarks, including nearly a 10\% LPIPS reduction on DAVIS and SNU-FILM, and an 8\% improvement on DAIN-HD.
\end{abstract}

\blfootnote{~~~\Letter: Corresponding author (zxwu@fudan.edu.cn).}

%% file: sec/1_intro.tex
\vspace{-3mm}
\section{Introduction}
\label{sec:intro}
Video frame interpolation is a classic computer vision task that aims to synthesize intermediate frames between given starting and ending frames. It has wide applications, including animation production~\cite{deepanimate}, generating slow motion videos~\cite{slomo, safa, videoinr}, novel view synthesis~\cite{LearningTo, deepstereo, blur} and videos compression~\cite{NCM, vidcom}.  

\begin{figure}[h]
  \centering
  \begin{tikzpicture}
        \node[anchor=south west, inner sep=0] (image) at (0,0) {
            \includegraphics[width=1.0\linewidth]{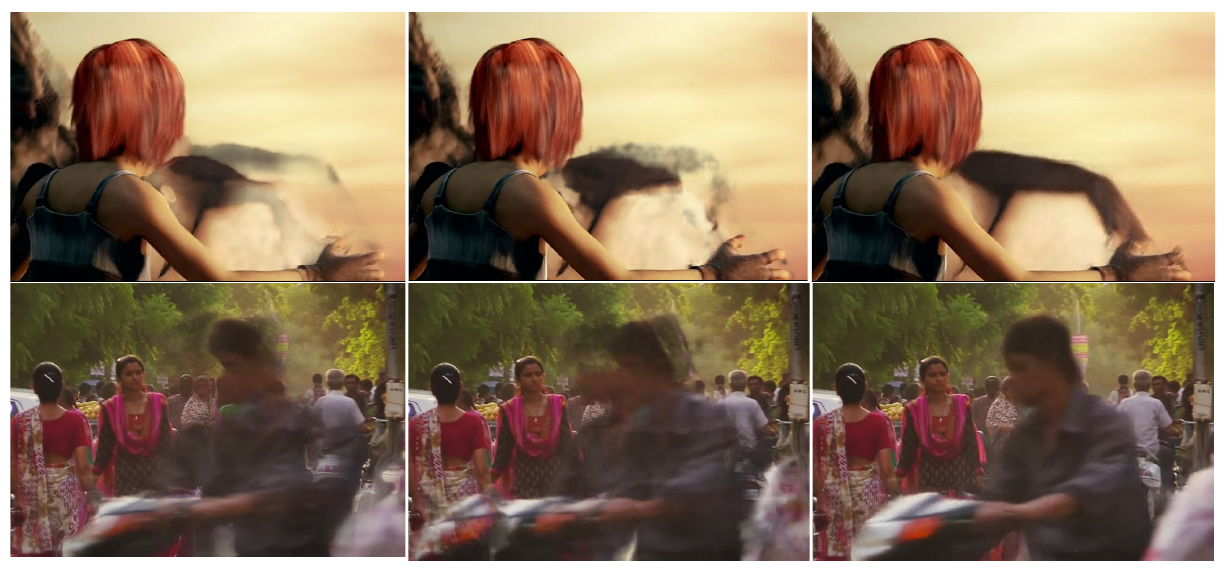}
        };
        \node[anchor=north west, color=black] at (0.5,0.0) {\fontsize{8}{10}\selectfont LDMVFI\cite{LDMVFI}};
        \node[anchor=north west, color=black] at (3.5,0.0) {\fontsize{8}{10}\selectfont LBBDM\cite{LBBDM}};
        \node[anchor=north west, color=black] at (6.5,0.0) {\fontsize{8}{10}\selectfont \textbf{Ours}};
    \end{tikzpicture}
  \vspace{-4ex}
  \caption{Comparisons between existing diffusion-based methods and our proposed \modelname~on large-motion scenarios.}
  \label{fig:diff_failure}
  \vspace{-1ex}
\end{figure}
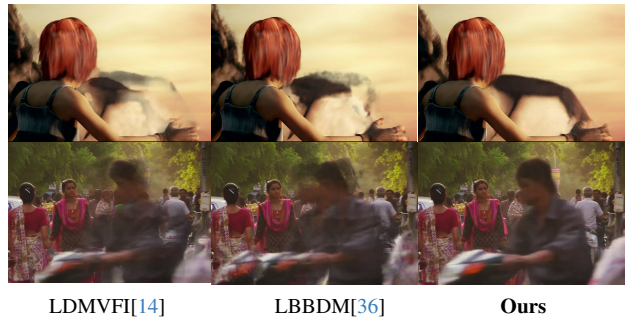

Most existing video frame interpolation methods~\cite{EMA, RIFE, BiFormer, VFIT, IFRNet, AMT, dav, SGMVFI, vfimamba} rely on estimating optical flow between the starting and ending frames to synthesize the intermediate frames. Recently, advances in diffusion models have inspired researchers to explore their application in video frame interpolation~\cite{MCVD, LDMVFI, LBBDM, madiff, VIDIM}. These diffusion-based approaches typically involve compressing the intermediate frames into latent representations using an encoder. Diffusion processes are then applied within the latent space to integrate information from both the starting and ending frames to predict the latent representation of the intermediate frame. Finally, a decoder reconstructs the latent representation into the desired intermediate frame. By directly 
 generating intermediate frames from latent representations, these models eliminate the need for explicit optical flow-based warping, offering a promising alternative to traditional methods.

\begin{figure}[t]
  \centering
    \begin{tikzpicture}
        \node[anchor=south west, inner sep=0] (image) at (0,0) {
            \includegraphics[width=1.0\linewidth]{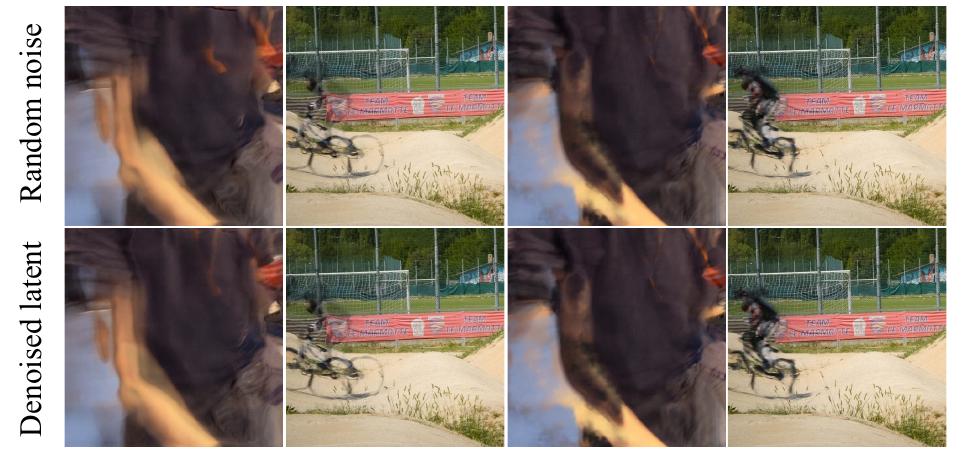}
        };
        \node[anchor=north west, color=black] at (1.7,0.0) {\fontsize{8}{10}\selectfont LDMVFI\cite{LDMVFI}};
        \node[anchor=north west, color=black] at (5.65,0.0) {\fontsize{8}{10}\selectfont LBBDM\cite{LBBDM}};
    \end{tikzpicture}
  \vspace{-4ex}
  \caption{The results of existing diffusion-based VFI methods generated from random noise and denoised latent. As is shown, the quality of the generated frame shows no significant difference.}
  \label{fig:noise_decode}
  \vspace{-1ex}
\end{figure}

However, despite their advantages, diffusion-based methods still often fail at accurately capturing large, complex motions. In scenes involving rapid, nonlinear movements, these approaches often face limitations in generating sharp, realistic frames, resulting in motion blur and temporal inconsistencies. For example, as shown in \Cref{fig:diff_failure}, current diffusion based methods struggle with handling large motions.
Furthermore, our analysis of current diffusion based methods reveals that the effect of diffusion on the quality of generated intermediate frames is surprisingly minimal. To illustrate, we present in \Cref{fig:noise_decode} a comparison between intermediate frames generated by directly decoding random noise versus decoding latent predicted by diffusion. As shown, the perceptual difference between the two is marginal, indicating a possible reason why these methods still struggle with complex motions.

To address this issue, in this paper, we aim to amplify the impact of diffusion in the generation process. We seek to accomplish this by focusing on the three primary areas where modifications are typically feasible: the input representation, model architecture, and training paradigm. By improving each of these essential components, we aim to deliver a more effective and resilient diffusion process.

Specifically, to enhance the quality of the latent representations input to the diffusion model, instead of using the usual 2D-convolution-based VAE structure, we first design an efficient transformer-based tokenizer that compresses the intermediate frames into a compact set of latent tokens. This is evidenced by a recent work~\cite{titok} that shows 1D sequences as latent representations learns more high-level and semantic-rich image information than prior 2D grid representations, especially at a compact latent space. 
Then, to better integrate the context information from the starting and ending frames, we incorporate temporal modules throughout the encoding-decoding process. 
Additionally, we propose a novel multi-scale feature fusion module that iteratively integrates large-scale tokens with small-scale tokens to capture fine-grained details across varying motion scales. Finally, given that real-world videos exhibit diverse resolutions and motion magnitudes, we further introduce a multi-resolution, multi-frame interval fine-tuning technique to handle these variations better.

To provide a better model architecture, we adopt a DiT-based diffusion model instead of the traditional U-Net approach, leveraging DiT’s enhanced capability to handle temporal dependencies, as well as its inherent support for token-based inputs and greater scalability compared to U-Net.
Furthermore, to improve the training paradigm, we propose a dual-stream context integration mechanism. Specifically, similar to the design of our tokenizer, we integrate temporal modules throughout the diffusion process to integrate the context information from the starting and ending frames, allowing the model to capture motion dynamics more effectively. Additionally, given that the motion magnitude between the starting and ending frames plays a critical role in the quality of generated frames, we introduce a start-end frame difference embedding as a conditioning input to the diffusion model. This embedding explicitly guides the generation process by embedding motion cues into the model.

Extensive experiments on popular benchmarks show the success of our approach. Specifically, our method achieves nearly a 10\% reduction in LPIPS (lower is better) compared to previous state-of-the-art diffusion models on DAVIS. On the more dynamic DAIN-HD benchmark, our method reduces LPIPS by 8\% compared to prior state-of-the-art methods specifically designed for large motion, as well as recent state-of-the-art methods that have performed well across multiple benchmarks. 
In summary, our main contributions include:

\begin{itemize}
    \item We introduce an enhanced diffusion-based method for high-quality video frame interpolation, which we dub \modelname, addressing the challenging problem of video frame interpolation with large motion.
    \item We design an efficient tokenizer for video frame interpolation, which enhances the quality of latent representations and validates its effectiveness through a series of ablation experiments.
    \item We propose a dual-stream context integration mechanism to incorporate context information from the starting and ending frames into the diffusion model, effectively enhancing the diffusion process.
    \item \modelname~achieves better results compared to prior state-of-the-art methods on large motion video frame interpolation benchmarks across generative modeling metrics.
\end{itemize}

%% file: sec/2_related_work.tex
\begin{figure*}[h]
\vspace{-2ex}
  \centering
  \includegraphics[width=1.0\linewidth]{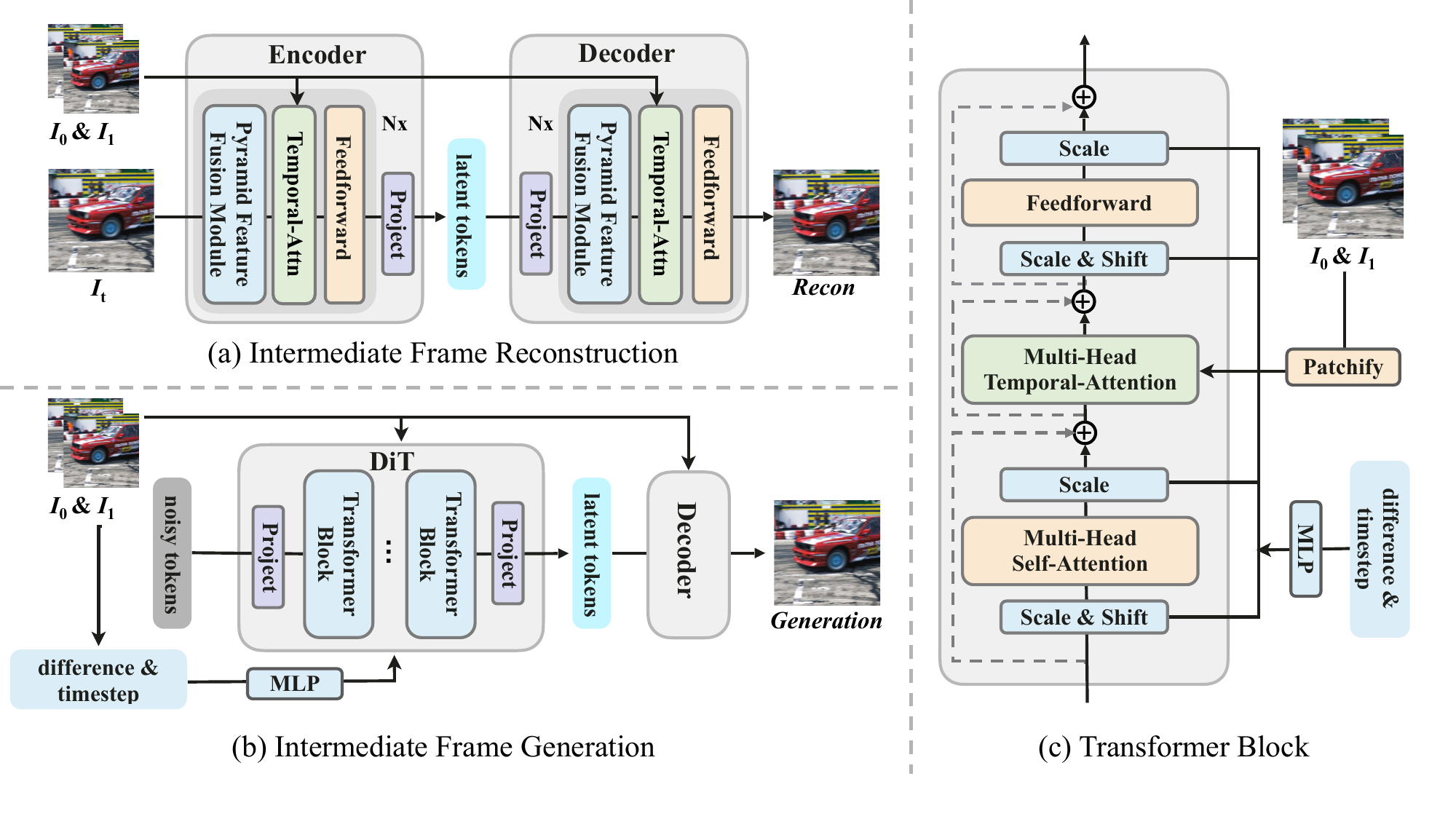}
  \vspace{-0.4in}
  \caption{\textbf{Illustration of intermediate frame reconstruction (a) and generation (b) with the transformer block (c). }First, we train a tokenizer by reconstructing the intermediate frames to obtain latent tokens with strong representational capabilities. Then, we train a diffusion transformer based on these latent tokens. During inference, we utilize the diffusion transformer to generate latent tokens from noise, which are then decoded into the intermediate frame by the tokenizer decoder. We inject the starting and ending frame information into the transformer block through temporal-attention and difference embedding.}
  \label{fig:pipeline}
  \vspace{-2ex}
\end{figure*}
\section{Related Work}
\label{sec:related_work}
\subsection{Video Frame Interpolation}
Most recent video frame interpolation methods focus on modeling the intermediate optical flow, then using forward warping~\cite{softmax} or backward warping~\cite{stmfnet, VFIFormer, IFRNet, RIFE, AMT} to synthesize the target frame. For example, SGM-VFI~\cite{SGMVFI} adopts a sparse global matching to refine the estimated flow. VFIMamba~\cite{vfimamba} utilizes a global receptive field to capture motion via state space models. 

Recent works have started to utilize diffusion models to handle the VFI task. For example, LDMVFI~\cite{LDMVFI} employs a latent diffusion model to generate the intermediate frame conditioned on given frames. LBBDM~\cite{LBBDM} utilizes a consecutive brownian bridge diffusion model to synthesize the target frame. MADIFF~\cite{madiff} uses diffusion models conditioned motion to generate the intermediate frame. VIDIM~\cite{VIDIM} uses cascaded diffusion models to generate target frames. However, despite their advantages, these diffusion-based methods still struggle with accurately capturing large, complex motions. In contrast, our \modelname~can effectively handle complex motions by improving the quality of the latent representations, modifying a better diffusion architecture, and enhancing the training paradigm.

\subsection{Image Tokenization}
Image tokenization has been widely used in deep learning, aiming to compress an image into a small number of informative tokens. Previous methods typically employ tokenizer to help large language models understand image information~\cite{blip2, visualinstuning, flamingo, llama2} or to guide generative models in image generation~\cite{ldm, pixart, sd3}, often without the need to reconstruct the original image from the compressed tokens. Recently, Titok~\cite{titok} explored the idea of compressing an image into a fixed number of tokens that can be used to reconstruct the original image. The authors show that in terms of obtaining better latent representations, 1D sequences are much more effective than 2D ones. Inspired by it, we design an efficient transformer-based tokenizer specifically for video frame interpolation tasks. 

\subsection{Diffusion Models}
\vspace{-1ex}
Denoising diffusion probabilistic models~\cite{ddpm} have gained considerable attraction due to their ability to produce high-quality, diverse outputs. These models generate data by reversing a diffusion process that progressively adds noise, allowing for fine control over the generation process.

Diffusion models have been applied to a range of tasks beyond traditional image generation. For image super-resolution, ~\cite{sr3, isr} demonstrates that diffusion models can produce high-fidelity results with better detail preservation than GANs. Similarly, for inpainting tasks, ~\cite{repaint, latentpaint} can seamlessly fill missing regions in images, often outperforming other generative methods in terms of texture and detail consistency. For video tasks, diffusion models have been used in video editing~\cite{magdiff, motioneditor, motionfollower} and video generation~\cite{svd, Latte, stableanimator}, where temporal consistency is critical. 

Moreover, the adaptability of diffusion models has led to the exploration of transformer-based diffusion backbones~\cite{dit, sit, sd3, sora, pixart}, which better handle both spatial and temporal dependencies, especially in tasks that require extensive contextual information. This development suggests that diffusion models, given their flexibility and scalability, could enable applications that require multi-scale reasoning and conditional information injection. Therefore, we employ the diffusion transformer as our generative model and make specific modifications to optimize it for VFI.

%% file: sec/3_method.tex
\section{Method}
\label{sec:method}
Our goal is to amplify the impact of the diffusion model in VFI task from the perspective of improving latent representation, modifying model architecture, and enhancing the training paradigm. To do so, we present the framework \textit{\textbf{\modelname}}~where we design a novel intermediate frame tokenizer based on the transformer architecture and a novel DiT based diffusion framework more suitable for video frame interpolation task.  \Cref{fig:pipeline} illustrates the pipeline. In the following sections, we analyze the technical details of how each of the aspects is implemented.

\subsection{Improving Latent Representation}
\label{sec:PFFM}
\noindent \textbf{Transformer-based Tokenizer.} Existing diffusion-based methods typically use a 2D-convolution-based VAE model to obtain the latent representations of intermediate frames. Inspired by TiTok \cite{titok}, which demonstrates that a sequence of 1D representation captures richer semantic information than a sequence of 2D representation, we design a transformer-based tokenizer that compresses the intermediate frame into a set of latent tokens. As shown in \Cref{fig:pipeline} (a), both our encoder and decoder consist of $N$ consecutive transformer blocks, where in addition to the regular self-attention and feed-forward module, each block also consists of a pyramid feature fusion module and temporal-attention modules (will be introduced next). Additionally, we employ a projection layer to reduce features into lower-dimensional tokens and then expand these tokens back to higher-dimensional features at the end and start of the encoder and decoder, respectively.

\vspace{1ex}

\begin{figure}[h]
  \centering
  \includegraphics[width=1.0\linewidth]{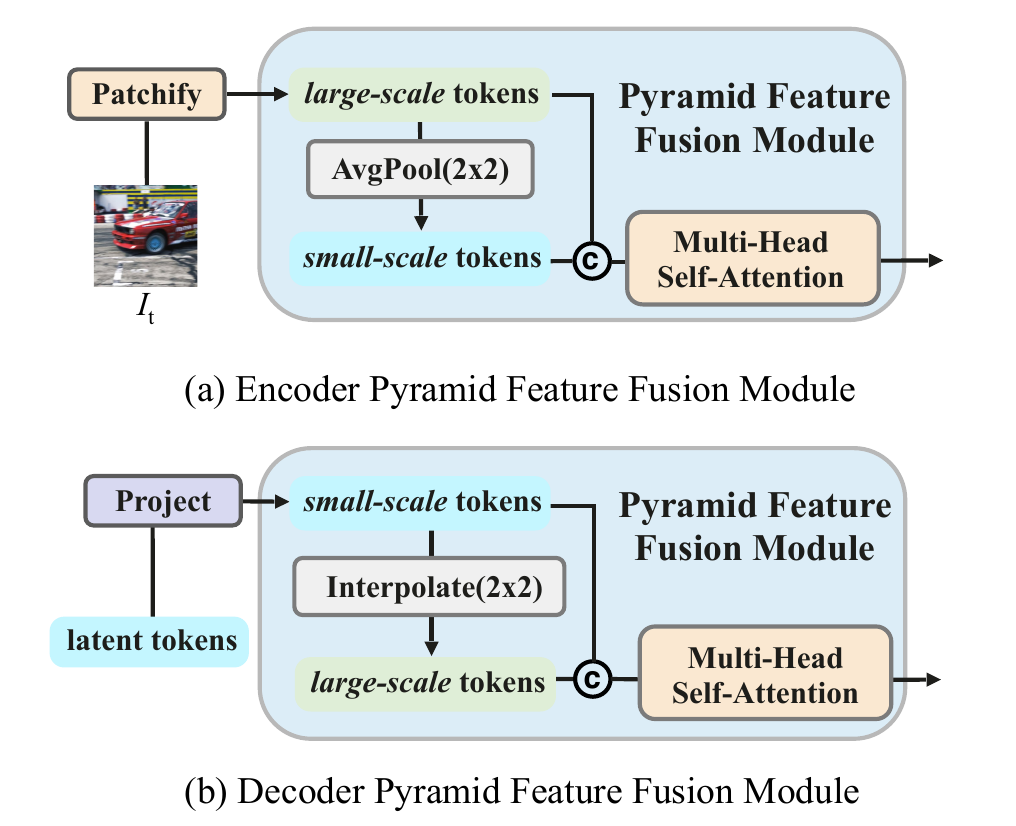}
  \vspace{-0.2in}
  \caption{Pyramid Feature Fusion Module design of encoder (a) and decoder (b).}
  \label{fig:PFFM}
\end{figure}

\noindent \textbf{Pyramid Feature Fusion Module.} Inspired by the capacity of multi-scale features to capture fine-grained details across varying motion scales~\cite{FILM, XVFI, EMA, SGMVFI}, we propose a Pyramid Feature Fusion Module to fuse features across tokens of different scales. Specifically, in the encoder, we first partition the input image into small patches to create a set of large-scale tokens. We then apply average pooling to generate a down-sampled set of small-scale tokens. Before passing these through the self-attention layer, we concatenate the large-scale and small-scale tokens, allowing self-attention to facilitate multi-scale feature fusion. In contrast, the decoder starts with a set of small-scale tokens. To align with the encoder’s feature scale, we interpolate these tokens to obtain the corresponding set of large-scale tokens. The difference is illustrated in \Cref{fig:PFFM}. After the self-attention layer processes the concatenated tokens, we selectively extract only the tokens at the positions corresponding to the small-scale features, passing them to subsequent layers for further processing.

\vspace{1ex}
\noindent \textbf{Temporal-Attention Module.}
Unlike traditional image compression approaches, video frame interpolation can benefit from the starting and ending frames by serving them as external guidance signals. To leverage this, we introduce temporal attention during the compression process, allowing the model to selectively retain relevant information based on token positions in the starting and ending frames. The goal is to filter out redundant information, thus producing latent tokens with improved representational capacity.
 
\vspace{1ex}

\noindent \textbf{Pipeline of a Transformer Block.} 
Formally, let $I_0, I_1, I_t$ be the starting, ending, and intermediate frame. We first apply the usual patch embedding to obtain $\Tilde{I_0}, \Tilde{I_1}, \Tilde{I_t^l} \in \mathbf{R}^{b \times m \times d}$ that represents patchified tokens. 

For simplicity, we illustrate only the encoder. In the encoder, average pooling is first performed on $\Tilde{I_t^l}$ to obtain $\Tilde{I_t^s} \in \mathbf{R}^{b \times n \times d}$ that represents the set of small-scale tokens.  Here \( b \) is the batch size, \( m \) and \( n \) are the number of tokens(\(m=4n\)), and \( d \) is the hidden dimension.

Then, the forward pass of a transformer block is:

\[
    \Tilde{I_t} = \text{Concat}(\Tilde{I_t^l}, \Tilde{I_t^s}) \in \mathbf{R}^{b \times (m+n) \times d},
\]
\[
 \Tilde{I_t^s} = \text{Self-Attention}(\Tilde{I_t})[:, m:, :] \in \mathbf{R}^{b \times n \times d},
\]
\[
\Tilde{I_0},\Tilde{I_1} = \text{Reshape}(\Tilde{I_0}, \Tilde{I_1}) \in \mathbf{R}^{b \times n \times 4 \times d},
\]
\[
\Tilde{I_t} = \text{Concat}(\Tilde{I_0}, \Tilde{I_t^s}, \Tilde{I_1}) \in \mathbf{R}^{b \times n  \times 9 \times d},
\]
\[
\Tilde{I_t^s} = \text{Temporal-Attention}(\Tilde{I_t})[:, :, 4, :] \in \mathbf{R}^{b \times n \times d},
\]
\[
\Tilde{I_t^s} = \text{FeedForward}(\Tilde{I_t^s}).
\]

\vspace{1ex}

\noindent \textbf{Multi-resolution and Multi-frame Interval Fine-tuning Technique. }In video frame interpolation tasks, videos are often of varying resolutions. However, vision transformers face the scalability challenge across different resolutions. Several approaches~\cite{resformer, vitar} have been proposed to address this limitation. Unlike previous methods that involve adding new modules, we introduce a simple yet effective multi-resolution fine-tuning technique. Additionally, we propose a multi-frame interval fine-tuning approach to guide \modelname~in handling starting and ending frames with diverse motion magnitudes. Specifically, we first train the model at a fixed, lower resolution (256 $\times$ 448) by cropping the original training instances with a fixed small interval (1 $\sim$ 5) between the start and end frames. After achieving satisfactory performance, we randomly select pairs of frames with varied interval lengths and resolutions with equal probability for further training, while the position embeddings are interpolated for different lengths and resolutions.

\vspace{1ex}
\noindent \textbf{Training objective} Following~\cite{ldm, vae}, our loss function of tokenizer comprises an L1 loss, a perceptual loss~\cite{lpips}, a patch-based adversarial loss~\cite{ganloss}, and a slight KL penalty enforcing a standard normal distribution on the learned latent space~\cite{vae} as follows:
\begin{equation}
\label{eq:loss_func}
    \mathcal{L}_{tok} = \lambda_{1}\mathcal{L}_1 + \lambda_{p}\mathcal{L}_p + \lambda_{G}\mathcal{L}_{G} + \lambda_{kl}\mathcal{L}_{kl},
\end{equation}
where $\lambda_{1}$, $\lambda_{p}$, $\lambda_{G}$ and $\lambda_{kl}$ are balancing coefficients of each loss.

\subsection{Modifying Diffusion Architecture}
Our latent tokens interact continuously with the starting and ending frames during compression, building representations with strong temporal alignment. However, U-Net-based diffusion models, with convolution backbones, struggle to capture temporal dynamics and are limited to 2D sequence inputs. In contrast, diffusion transformers can seamlessly integrate temporal information via attention and handle 1D latent representations effectively. Therefore, we select the diffusion transformer, DiT~\cite{dit}, as the backbone of our generative model. 

In DiT, the backbone is composed of a series of transformer blocks, maintaining token scale throughout the network. This design effectively prevents information loss that often occurs due to downsampling in U-Net architectures. Each transformer block’s attention mechanism inherently provides a global receptive field, which captures large intra-frame correlation more effectively than the local receptive field of convolution layers. Moreover, the diffusion transformer uses adaptive Layer Normalization (adaLN) to modulate the input and output at each layer, facilitating efficient conditional information injection. Specifically, the modulation process is formulated as:
\[
(\alpha_1, \beta_1, \gamma_1, \alpha_2, \beta_2, \gamma_2)= \text{AdaLN}(c),
\]
\vspace{-3ex}
\[
x = x + \alpha_1\text{SelfAttention}(x\times\gamma_1 + \beta_1),
\]
\vspace{-3ex}
\[
x = x + \alpha_2\text{FeedForward}(x\times\gamma_2 + \beta_2),
\]
where $c$ is the conditional input and $x$ is the feature map from previous layers.

\subsection{Enhancing Training Paradigm}
\label{sec:dual_stream_condition}
As shown in the \Cref{fig:pipeline} (c), our diffusion transformer builds on the original DiT by introducing a dual-stream context integration mechanism to effectively incorporate context information from the starting and ending frames: one for temporal context and the other for difference context.

\vspace{1ex}
\noindent \textbf{Temporal Context Integration.} Recent works~\cite{sd3, pixart, Latte, sora} indicate that the original DiT’s method of incorporating label and timestep information through adaLN has limited capacity for embedding detailed contextual information. For video frame interpolation, it is essential to enhance the model’s ability to capture temporal dynamics and integrate information from the starting and ending frames into the diffusion process. To achieve this, we introduce a temporal attention layer after each self-attention layer similar to the ones in the diffusion transformer blocks of the tokenizer, enabling effective integration of complete start-end frame information throughout the diffusion process.

\vspace{1ex}
\noindent \textbf{Difference Context Integration.} In video frame interpolation, the diffusion generation process differs significantly from that of standard image generation. Here, our latent tokens represent intermediate frames, which are intermediate states of the starting and ending frames, rather than static 2D images. These tokens are significantly influenced by the differences between the starting and ending frames, and any variations in these differences can affect the diffusion transformer's output. To address this, we introduce explicit integration of start-end frame difference context, guiding the diffusion process and enhancing the model's capacity to adapt to fluctuations between the initial and final frames.

Specifically, we calculate the mean and standard-deviation of the cosine similarity for all the start-end frame pairs in the training set. During training, the cosine similarity of each start-end frame pair is normalized by subtracting the mean and dividing by the standard-deviation to obtain a difference context. We then use an MLP to convert this difference context into a difference embedding, which is added to the timestep embedding and used as the conditional input for adaLN, as is shown in \Cref{fig:pipeline} (b). 

\vspace{1ex}
\noindent \textbf{Training Objective.} 
Following the training strategy of ~\cite{sd3}, we define the forward process as straight paths between the data distribution $x_0 \sim p(x)$ and a standard normal distribution $\eps \sim \mathcal{N}(0, I)$:
\begin{align}
    x_t = (1 - t) x_0 + t \eps, t \in [0, 1],
\end{align}
and the diffusion model predicts the following velocity field that maps the noise distribution to the data distribution:
\begin{equation}
  \label{eq:ode}
  dx_t = u_t(x_t, t)\,dt,
\end{equation}
with the following loss function:
\begin{align}
   \mathcal{L}_{dit} =  \mathbb{E}_{t, p_t(z)} || v_{\Theta}(z, t) - u_t(z) ||_2^2.
\end{align}


%% file: sec/4_experiments.tex
\begin{table*}[h]
\centering
\resizebox{\textwidth}{!}{
\begin{tabular}{lcccccccccc}
\toprule
\multirow{2}{*}{} & \multicolumn{3}{c}{\textbf{DAVIS}} & \multicolumn{3}{c}{\textbf{DAIN-HD544p}} & \multicolumn{2}{c}{\textbf{SNU-FILM (Easy)}} & \multicolumn{2}{c}{\textbf{SNU-FILM (Extreme)}} \\ 
\cmidrule(lr){2-4} \cmidrule(lr){5-7} \cmidrule(lr){8-9} \cmidrule(lr){10-11} 
                  & LPIPS$\downarrow$ & FloLPIPS$\downarrow$ & RT(s) & LPIPS$\downarrow$ & FloLPIPS$\downarrow$ & RT(s) & LPIPS$\downarrow$ & FloLPIPS$\downarrow$ & LPIPS$\downarrow$ & FloLPIPS$\downarrow$ \\ 
\midrule
VFIFormer~\cite{VFIFormer} & 0.1370 & 0.1905 & 1.178 & 0.1604 & 0.2764 & 2.319 & 0.0176 & 0.0290 & 0.1463 & 0.2361 \\
ST-MFNet~\cite{stmfnet}    & 0.1232 & 0.1674 & 0.364 & 0.1577 & 0.2507 & 0.730 & 0.0210 & 0.0336 & 0.1302 & 0.2192 \\
AMT~\cite{AMT}             & 0.1091 & 0.1523 & \textbf{0.122} & 0.1479 & 0.2697 & {\ul 0.254} & 0.0191 & 0.0309 & 0.1205 & 0.2054 \\
SGM-VFI~\cite{SGMVFI}      & 0.1140 & 0.1571 & 0.136 & {\ul 0.1423} & 0.2471 & 0.380 & 0.0178 & 0.0292 & 0.1205 & 0.1936 \\
VFIMamba~\cite{vfimamba}  & 0.1084 & 0.1486 & 0.230 & 0.1426 & 0.2471 & 0.380 & 0.0179 & 0.0299 & 0.1154 & 0.1892 \\
LDMVFI~\cite{LDMVFI}      & 0.1095 & 0.1497 & 7.001 & 0.1490 & 0.2378 & 8.304 & 0.0145 & 0.0207 & 0.1227 & 0.1894 \\
LBBDM~\cite{LBBDM}        & {\ul 0.0963} & {\ul 0.1313} & 1.689 & 0.1471 & {\ul 0.2319} & 2.420 & \textbf{0.0120} & \textbf{0.0162} & {\ul 0.1101} & {\ul 0.1696} \\
\midrule
\textbf{Ours} & \textbf{0.0874} & \textbf{0.1201} & {\ul 0.130} & \textbf{0.1321} & \textbf{0.2184} & \textbf{0.250} & {\ul 0.0129} & {\ul 0.0179} & \textbf{0.0986} & \textbf{0.1653} \\
\bottomrule
\end{tabular}
}
\caption{Quantitative evaluation across different challenging benchmarks. The best results are in \textbf{bold}, and the second best are {\ul underlined}. Running Time (RT) refers to the time required to interpolate one frame. All experiments were tested on an NVIDIA V100-32G GPU.}
\label{tab:comparisons}
\end{table*}

\section{Experiments}
\label{sec:experiments}

\subsection{Implementation Detail}
\noindent \textbf{Datasets. }Similar to previous diffusion-based works~\cite{LDMVFI, LBBDM, madiff, VIDIM}, we employ a large-scale dataset LAVIB~\cite{lavib} for training, aiming to effectively assess the scalability of diffusion models. LAVIB contains 284,484 clips with a resolution of 1296x1296, each clip has 60 consecutive frames. To test the effectiveness of our method on large and complex motion tasks, we select commonly recognized large motion datasets DAVIS~\cite{davis}, DAIN-HD544p~\cite{hd}, and SNU-FILM~\cite{snufilm} as our testing datasets. DAVIS is a widely used benchmark for large motions, containing 2,849 triplets at a resolution of 480x854. DAIN-HD544p is a more dynamic dataset, featuring 124 triplets at a resolution of 544x1280. SNU-FILM is a dataset comprising subsets with varying motion magnitudes, where each subset contains 309 triplets at multiple resolutions.

\noindent \textbf{Hyperparameters. }To train the designed tokenizer and diffusion model, each instance is cropped to patches with a batch size of 256, without any data augmentation. We use AdamW~\cite{adam} as the optimizer with $\beta_1=0.9$, $\beta_2=0.99$, and weight decay $1e^{-4}$. Cosine annealing is used for learning rate decay from $1.0e^{-4}$ to $1.0e^{-8}$. In the multi-resolution and multi-frame interval fine-tuning stage, the batch size is reduced to 64 and the learning rate decays from $1.0e^{-5}$ to $1.25e^{-8}$. The weights for \Cref{eq:loss_func} are set to $1.0, 1.0, 0.5, 1.0e^{-6}$ for $\lambda_{1}, \lambda_{p},\lambda_{G}$ and $\lambda_{kl}$, respectively. The tokenizer uses 4 transformer blocks for both encoder and decoder, while the diffusion transformer uses 12 blocks, with a hidden dimension of 768 for all blocks.

\noindent \textbf{Evaluation metrics. }
Similar to previous diffusion-based work~\cite{VIDIM, LDMVFI, LBBDM, madiff}, we report perceptual metrics LPIPS~\cite{lpips} and FloLPIPS~\cite{flolpips} instead of reconstruction-based metrics like peak-signal-to-noise-ratio (PSNR) and structural similarity (SSIM)~\cite{ssim}. It is well-known that generative models are not expected to achieve the best scores in reconstruction-based metrics ~\cite{ddpmsr, nvs}. In fact, prior work~\cite{fsrnet, prsr, pulse, isr} has consistently shown that blurrier images often score higher in reconstruction metrics, despite being rated as worse by human observers.

\vspace{-1ex}
\subsection{Comparisons with Previous Works}
\vspace{-1ex}
To comprehensively evaluate the effectiveness of our model in large motion video frame interpolation tasks, we compare it with both traditional optical flow-based methods and recent diffusion-based approaches. Specifically, we include VFIFormer~\cite{VFIFormer}, ST-MFNet~\cite{stmfnet}, AMT~\cite{AMT}, SGM-VFI~\cite{SGMVFI}, VFIMamba~\cite{vfimamba}, LDMVFI~\cite{LDMVFI}, and LBBDM~\cite{LBBDM} for comparison.

\begin{figure*}[h]
    \centering
    \includegraphics[width=1.0\linewidth]{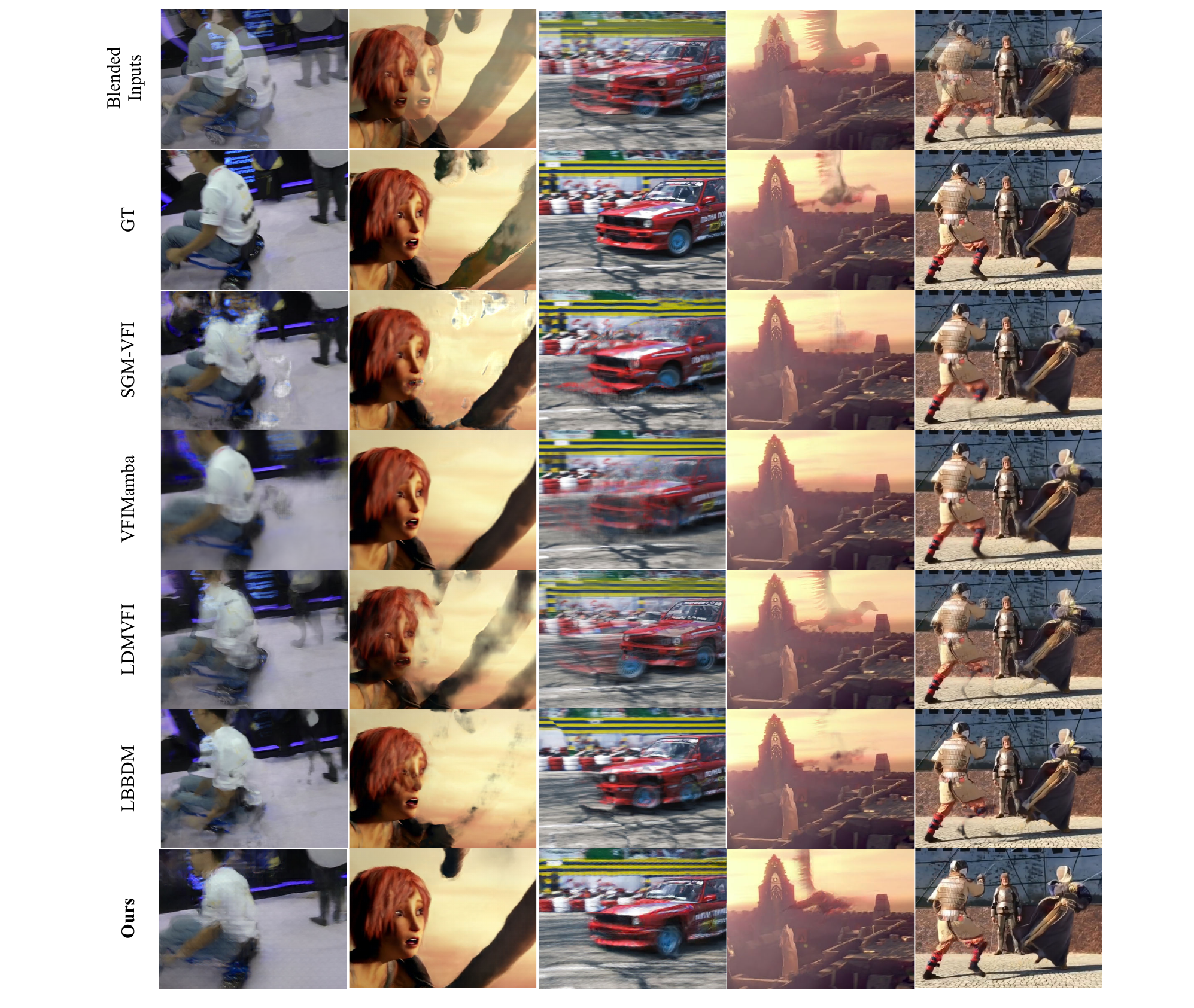}
    \vspace{-0.2in}
    \caption{Visual comparison with different methods, examples selected from DAVIS, DAIN-HD544p and SNU-FILM. Ours outperforms previous methods in both capturing the motion of multiple objects and modeling fast, nonlinear motions.}
    \label{fig:visualization}
\end{figure*}

The results on DAVIS, DAIN-HD544p, SNU-FILM are shown in \cref{tab:comparisons}. For DAVIS, it is obvious that \modelname~outperforms all other methods, achieving an LPIPS of 0.0874 and a FloLPIPS of 0.1201. To the best of our knowledge, these are the best perceptual results reported on DAVIS so far. For DAIN-HD544P, \modelname~also achieves the best results on this more dynamic dataset, with an LPIPS of 0.1321 and a FloLPIPS of 0.2184. Compared to SGM-VFI, which is specifically designed for large motion, and VFIMamba, which has shown state-of-the-art performance across various benchmarks, our LPIPS improves by 8\%. For SNU-FILM, \modelname~achieves the best LPIPS of 0.0986 in the extreme motion subset, and also obtains the second-best LPIPS of 0.0129 in the easy motion subset, slightly below the best LPIPS of 0.0120. Notably, we find that only 2 denoising steps are sufficient to achieve high-quality generation (as shown in \cref{tab:denoising_steps} and \cref{fig:eden_noise}). Consequently, \modelname~is significantly faster in inference time compared to previous diffusion-based methods, even outpacing some optical flow-based methods, and is only slower than the fastest methods.

\vspace{-1ex}
\subsection{Visualization}
\vspace{-1ex}
\cref{fig:visualization} presents a visual comparison between \modelname~and previous methods. In the first example, where the left side shows a moving person and the right side shows the corresponding lower body movement, as well as in the second example, where a moving face and an animal's tail are depicted, previous methods are limited to capturing the motion of a single object. In contrast, our approach effectively models the motion of multiple objects within the same frame. Notably, in scenarios involving fast or nonlinear motion -- such as in the third example with the car, the fourth example with the bird, and the fifth example with the human leg and sword -- previous methods struggle to accurately capture the motion. Our method, however, successfully tracks the motion, according to the changes between the starting and ending frames.

\vspace{-1ex}
\subsection{Ablation Study}
\vspace{-1ex}
\begin{table}[h]
\centering
\begin{tabular}{@{}lcc@{}}  
\toprule
\multirow{2}{*}{Placement of Attention} & \multicolumn{2}{c}{VIMEO} \\ \cmidrule(lr){2-3}
 & LPIPS$\downarrow$ & FloLPIPS$\downarrow$ \\ \midrule
None block         & 0.1731  & 0.0473  \\
Encoder blocks     & 0.0773  & 0.0311  \\
Decoder blocks     & 0.0190  & 0.0121  \\
All blocks & \textbf{0.0150}  & \textbf{0.0103}  \\ \bottomrule
\end{tabular}
\vspace{-1ex}
\caption{Results of incorporating temporal attention in different blocks on the VIMEO~\cite{vimeo} dataset. All models are trained for 200k steps at a resolution of 256x448.}
\label{tab:attention_block}
\vspace{-4ex}
\end{table}

\label{sec:ablation}
\noindent \textbf{Placement of Temporal Attention.} We evaluate the effects of incorporating temporal attention in different transformer blocks of the intermediate frame tokenizer. The results, shown in \cref{tab:attention_block}, clearly indicate that incorporating temporal attention in both the encoder and decoder yields the best performance. This outcome aligns with our hypothesis that integrating temporal dependencies across both encoding and decoding stages enables a more comprehensive capture of motion dynamics, leading to improved frame interpolation quality.

\vspace{1ex}
\noindent \textbf{Type of Incorporated Attention.} We also conduct a comparative analysis of cross-attention and temporal attention, with results summarized in \cref{tab:attention_type}. On the DAVIS dataset, both methods demonstrate similar performance, showing no substantial differences. However, on DAIN-HD544p, temporal attention yields a significant performance boost over cross-attention. This contrast implies that, for resolutions encountered during training, cross-attention and temporal attention perform comparably. Yet, when evaluated on previously unseen resolutions, cross-attention struggles to maintain performance, unlike temporal attention. This observed discrepancy may be due to temporal attention’s ability to implicitly align corresponding spatial positions across frames, enhancing robustness to resolution shifts.

\begin{table}[h]
\vspace{-1ex}
\centering
\resizebox{\columnwidth}{!}{
\begin{tabular}{ccccc}
\toprule
\multirow{2}{*}{Attention Type} & \multicolumn{2}{c}{DAVIS} & \multicolumn{2}{c}{DAIN-HD544p} \\ 
\cmidrule(lr){2-3} \cmidrule(lr){4-5}
& LPIPS $\downarrow$ & FloLPIPS $\downarrow$ & LPIPS $\downarrow$ & FloLPIPS $\downarrow$ \\ 
\midrule
Cross-Attn & 0.0547 & 0.0689 & 0.0718 & 0.0971 \\
Temporal-Attn & 0.0548 & 0.0695 & \textbf{0.0515} & \textbf{0.0665} \\ 
\bottomrule
\end{tabular}
}
\vspace{-1ex}
\caption{Results of cross-attention and temporal-attention on different resolution datasets: DAVIS and DAIN-HD544p. All models are trained for 200k steps at a resolution of 256x448, followed by an additional 50k steps of multi-resolution fine-tuning (includes resolution of 480x854 but excludes 544x1280).}
\label{tab:attention_type}
\vspace{-2ex}
\end{table}

\noindent \textbf{Latent Dimension.} Recent advancements in image generation~\cite{emu, sd3} suggest that increasing the latent dimension significantly improves the generative performance of diffusion models. However, increasing the latent dimension also heightens demands on the diffusion model's generative capacity. We assess tokenizer’s performance across different latent dimensions and evaluate these dimensions in combination with DiT in \cref{tab:dit_latents}. Notably, while the 24-dimensional tokenizer exhibits superior reconstruction capabilities, its combination with DiT underperforms compared to the 16-dimensional tokenizer, likely due to DiT’s limited generative capacity. Considering computational efficiency, we ultimately adopt the 16-dimensional tokenizer and provide additional experiments in \cref{tab:pffm_dit}.

\begin{table}[h]
\vspace{-1ex}
\centering
\resizebox{0.8\columnwidth}{!}{
\begin{tabular}{cccc}
\toprule
\multirow{2}{*}{} & \multirow{2}{*}{\begin{tabular}[c]{@{}c@{}}Latent \\ Dimension\end{tabular}} & \multicolumn{2}{c}{DAIN-HD544p} \\ 
\cmidrule(lr){3-4}
& & LPIPS $\downarrow$ & FloLPIPS $\downarrow$ \\ 
\midrule
\multirow{3}{*}{Tokenizer} & 4 & 0.0816 & 0.1236 \\
& 16 & \underline{0.0576} & \underline{0.0758} \\
& 24 & \textbf{0.0515} & \textbf{0.0665} \\ 
\midrule
\multirow{3}{*}{\begin{tabular}[c]{@{}c@{}}Tokenizer + \\ Diffusion\end{tabular}} & 4 & 0.2391 & 0.3099 \\
& 16 & \textbf{0.1538} & \textbf{0.2363} \\
& 24 & \underline{0.1641} & \underline{0.2510} \\ 
\bottomrule
\end{tabular}}
\vspace{-1ex}
\caption{Results of different latent dimensions on DAIN-HD544p. All models are trained for 200k steps at a resolution of 256x448, followed by 50k steps of multi-resolution fine-tuning.}
\label{tab:dit_latents}
\vspace{-2ex}
\end{table}

\noindent \textbf{Pyramid Feature Fusion Module. }To verify the effectiveness of our pyramid feature fusion module, we study the performance of tokenizer after removing pyramid feature fusion module in \cref{tab:PFFM_ab}. Specifically, we remove the concatenation of different scale tokens and directly use large-scale or small-scale tokens separately. Clearly, the performance of tokenizer is significantly worse without the pyramid feature fusion module.

\begin{table}[ht]
\vspace{-1ex}
\centering
\resizebox{\columnwidth}{!}{
\begin{tabular}{@{}lccccc@{}}
\toprule
\multirow{2}{*}{} & \multirow{2}{*}{PFFM} & \multicolumn{2}{c}{DAVIS} & \multicolumn{2}{c}{DAIN-HD544p} \\ \cmidrule(lr){3-4} \cmidrule(lr){5-6}
                  &                       & LPIPS$\downarrow$ & FloLPIPS$\downarrow$ & LPIPS$\downarrow$ & FloLPIPS$\downarrow$ \\ \midrule
Tokenizer         & $\times$              & 0.0564 & 0.0799 & 0.0596 & 0.0926 \\
Tokenizer         & $\surd$               & \textbf{0.0511} & \textbf{0.0669} & \textbf{0.0428} & \textbf{0.0626} \\ \bottomrule
\end{tabular}
}
\vspace{-1ex}
\caption{Results of our tokenizer with or without Pyramid Feature Fusion Module.}
\label{tab:PFFM_ab}
\vspace{-1ex}
\end{table}

\vspace{-2ex}
\noindent \textbf{Difference Context Integration.} As stated in \Cref{sec:dual_stream_condition}, our design of the diffusion transformer is guided by a difference embedding between the starting and ending frames. To verify the effectiveness of this strategy, we test the results with and without the difference embedding on the DAVIS and DAIN-HD544p datasets. The results in \cref{tab:difference_ab} demonstrate the significant impact of the difference context integration on the generative capability of the diffusion model.

\begin{table}[h]
\vspace{-1ex}
\centering
\resizebox{\columnwidth}{!}{  
\begin{tabular}{@{}cccccc@{}}  
\toprule
\multirow{2}{*}{}         & \multirow{2}{*}{\begin{tabular}[c]{@{}c@{}}Difference \\ Embedding\end{tabular}} & \multicolumn{2}{c}{DAVIS} & \multicolumn{2}{c}{DAIN-HD544p} \\ \cmidrule(lr){3-4} \cmidrule(lr){5-6}
                          &                                                                                  & LPIPS$\downarrow$      & FloLPIPS$\downarrow$     & LPIPS$\downarrow$       & FloLPIPS$\downarrow$     \\ \midrule
\text{\modelname}          & $\times$                                                                         & 0.0976     & 0.1327       & 0.1425      & 0.2376       \\
\text{\modelname}          & $\surd$                                                                          & \textbf{0.0874}     & \textbf{0.1201}       & \textbf{0.1321}      & \textbf{0.2184}       \\ \bottomrule
\end{tabular}
}
\vspace{-1ex}
\caption{Results of training the diffusion model with or without difference context integration.}
\label{tab:difference_ab}
\end{table}

\vspace{-1ex}

%% file: sec/5_conclusion.tex
\vspace{-2ex}
\section{Conclusion}
\vspace{-1ex}
\label{sec:conclusion}
In this paper, we introduced \modelname, an enhanced diffusion-based method designed to tackle the challenges of large motion in video frame interpolation. Our framework employed a transformer-based tokenizer to compress intermediate frames into compact tokens, enhancing latent representations for the diffusion process. To address multi-scale motion, we incorporated a pyramid feature fusion module and introduced multi-resolution and multi-frame interval fine-tuning to adapt the model to varying motion magnitudes and resolutions. By utilizing a diffusion transformer with temporal attention and a start-end frame difference embedding, \modelname~captured complex motion dynamics more effectively. Extensive experiments demonstrated that \modelname~achieved state-of-the-art performances on large-motion video benchmarks while also reducing computational costs. Despite these improvements, our method still suffers from blurring when handling fast changes in fine details (e.g., text). Discussions on further details are included in appendix.

\vspace{-2ex}
\paragraph{\noindent Acknowledgment.} This work is supported by NSFC under Grant No.62472098.

%% file: sec/X_suppl.tex
\clearpage
\setcounter{page}{1}
\appendix
\renewcommand\thefigure{\Alph{section}\arabic{figure}}
\renewcommand\thetable{\Alph{section}\arabic{table}}

\section{Appendix}
\subsection{Tokenizer Scalability}
\label{sec:scalability}
\begin{figure}[h]
    \centering
    \includegraphics[width=1.0\linewidth]{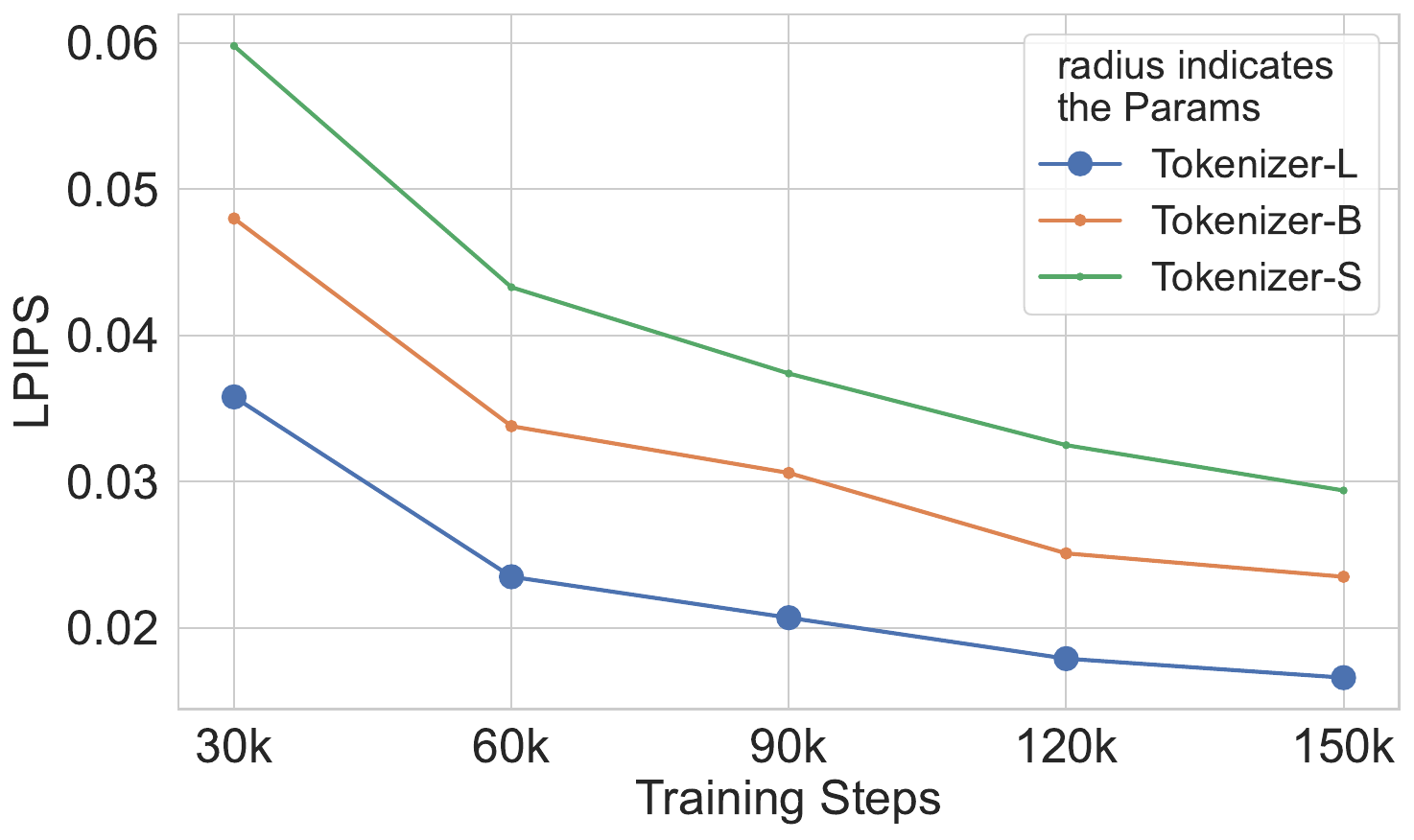}
    \vspace{-2ex}
    \caption{Scaling the tokenizer improves LPIPS at all stages of training. Tokenizer-S, Tokenizer-B, and Tokenizer-L have hidden dimensions of 384, 512, and 768, with parameter counts of 20.11M, 35.01M, and 77.35M, respectively.}
    \label{fig:model_size_comp}
\end{figure}

We briefly validate the scalability of our tokenizer in \cref{fig:model_size_comp} by leveraging different hidden dimensions of 384, 512 and 768. As is shown, increasing the hidden dimension improves the performance at all stages of training, demonstrating the scalability of our approach. We ultimately select Tokenizer-L for our final tokenizer.

\vspace{-4ex}
\subsection{Diffusion Denoising Steps}
\label{sec:denoising}
\begin{figure}[h]
    \centering
    \vspace{-2ex}
    \includegraphics[width=1.0\linewidth]{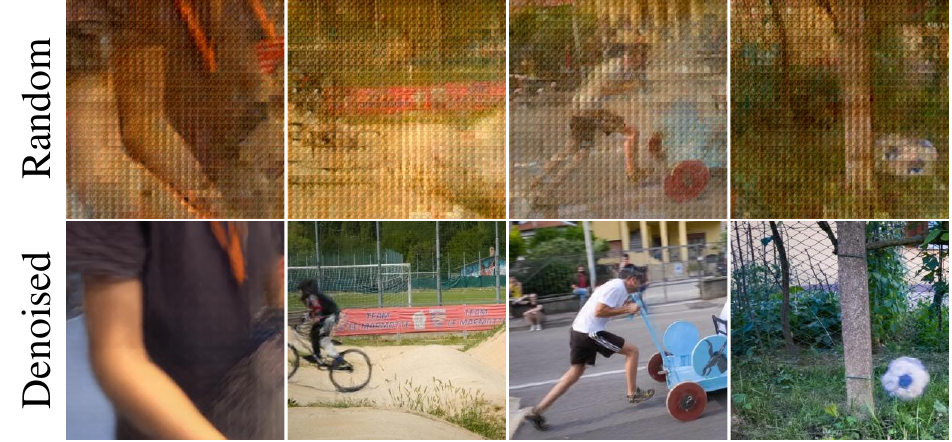}
    \vspace{-2ex}
    \caption{The results of EDEN generated from random noise and denoised latent.}
    \label{fig:eden_noise}
\vspace{-4ex}
\end{figure}

We evaluate the impact of the number of diffusion denoising steps on the interpolation performance of our model using DAVIS, with the visualization results in \cref{fig:eden_noise} and the quantitative results presented in \cref{tab:denoising_steps}. As observed, using random noise fails to capture accurate intermediate frame motion. Additionally, the model achieves satisfactory performance with just two denoising steps. Increasing the number of steps further significantly extends the denoising time while yielding limited improvement. Therefore, we ultimately select two denoising steps as the optimal choice. 
\begin{table}[ht]
\centering
\begin{tabular}{cccc}
\toprule
\multirow{2}{*}{Denoising Steps} & \multicolumn{3}{c}{DAVIS} \\ 
\cmidrule(l){2-4} 
 & LPIPS $\downarrow$ & FloLPIPS $\downarrow$ & RT (s) \\ 
\midrule
0  & 0.7882 & 0.7907 & 0.06 \\
1  & 0.0892 & 0.1216 & 0.09 \\
2  & 0.0874 & 0.1201 & 0.12 \\
5  & 0.0877 & 0.1201 & 0.28 \\
20 & 0.0880 & 0.1204 & 1.17 \\
50 & 0.0874 & 0.1200 & 2.81 \\ 
\bottomrule
\end{tabular}
\vspace{-1ex}
\caption{Performance comparison across different denoising steps on the DAVIS.}
\label{tab:denoising_steps}
\end{table}

\subsection{1D Tokenizer vs 2D VAE}
The ablation results of Transformer-based tokenizer and CNN-based VAE are summarized in \cref{tab:ab_ae}. As is shown, our tokenizer significantly outperforms 2D VAE of LBBDM in reconstructing intermediate frames.
\begin{table}[ht]
\centering
\resizebox{1.0\linewidth}{!}{\begin{tabular}{@{}lcccc@{}}
\toprule
                & PSNR$\uparrow$ & SSIM$\uparrow$ & LPIPS$\downarrow$ & FloLPIPS$\downarrow$ \\ 
\midrule
LBBDM VAE       & 26.64          & 0.945         & 0.1471            & 0.2319               \\ 
Our Tokenizer   & \textbf{34.93} & \textbf{0.978} & \textbf{0.0428}   & \textbf{0.0626}      \\ 
\bottomrule
\end{tabular}}
\vspace{-1ex}
\caption{Performance comparison on the DAIN-HD544p dataset.}
\label{tab:ab_ae}
\end{table}

\vspace{-4ex}
\subsection{Increasing interval of input frames}
We evaluate the performance of LBBDM and EDEN on DAIN-HD544p under increasing frame intervals, with the results summarized in \cref{tab:tdr}. As is shown, EDEN's performance in fitting intermediate frames still consistently outperforms LBBDM.
\begin{table}[ht]
\centering
\resizebox{1.0\linewidth}{!}{\begin{tabular}{@{}lcccccc@{}}
\toprule
 & Method & PSNR$\uparrow$     & SSIM$\uparrow$     & LPIPS$\downarrow$     & FloLPIPS$\downarrow$ \\ \midrule
\multirow{2}{*}{2x} 
                & LBBDM  & 26.64             & 0.945             & 0.1471               & 0.2319               \\
                & EDEN   & \textbf{26.85}    & \textbf{0.945}    & \textbf{0.1321}      & \textbf{0.2184}      \\ \midrule
\multirow{2}{*}{4x} 
                & LBBDM  & 19.55             & 0.855             & 0.2665               & 0.3378               \\
                & EDEN   & \textbf{20.91}    & \textbf{0.869}    & \textbf{0.2295}      & \textbf{0.3064}      \\ \midrule
\multirow{2}{*}{8x} 
                & LBBDM  & 16.09            & 0.764             & 0.4026               & 0.4376               \\
                & EDEN   & \textbf{17.70}   & \textbf{0.795}    & \textbf{0.3328}      & \textbf{0.3828}      \\ \bottomrule
\end{tabular}}
\vspace{-1ex}
\caption{Comparison of LBBDM and EDEN under different temporally downsample rates on the DAIN-HD544p dataset.}
\label{tab:tdr}
\end{table}

\vspace{-4ex}
\subsection{Multi-fine-tuning Techniques}
We conduct such ablation study on the DAIN-HD544p dataset. The results in \cref{tab:ad_ft} show a clear improvement in the tokenizer's performance after fine-tuning.
\begin{table}[h]
\centering
\resizebox{1.0\linewidth}{!}{\begin{tabular}{lcccc}
\toprule
         & PSNR$\uparrow$ & SSIM$\uparrow$ & LPIPS$\downarrow$ & FloLPIPS$\downarrow$ \\ 
\midrule
\textit{wo/fine-tuning} & 23.55         & 0.901         & 0.3314         & 0.3616            \\ 
\textit{w/fine-tuning}  & \textbf{34.93} & \textbf{0.978} & \textbf{0.0428} & \textbf{0.0626}  \\ 
\bottomrule
\end{tabular}}
\vspace{-1ex}
\caption{Performance comparison with and without fine-tuning.}
\label{tab:ad_ft}
\vspace{-2ex}
\end{table}

\vspace{-4ex}
\subsection{Ablation studies of tokenizer dimension}
An autoencoder’s reconstruction quality directly constrains the achievable image quality in latent diffusion models. The contradiction in \cref{tab:dit_latents} of the main paper arises because we used only DiT-B in EDEN due to computational constraints. However, higher-dimensional tokenizers (24) require larger DiTs to fully capture distribution transitions. 

To further clarify, we also provide ablation results of the tokenizer combined with DiT on DAIN-HD544p, training for 200k steps, as shown in \cref{tab:pffm_dit}. Clearly, the interpolation performance of the tokenizer aligns well with its reconstruction capability (as shown in the main paper) when integrated with DiT.
\begin{table}[ht]
\centering
\begin{tabular}{@{}llcc@{}}
\toprule
& PFFM                      & LPIPS$\downarrow$ & FloLPIPS$\downarrow$ \\ \midrule
Tokenizer + DiT-B         & $\times$              & 0.1528 & 0.2531 \\
Tokenizer + DiT-B        & $\surd$               & \textbf{0.1497} & \textbf{0.2503} \\ \bottomrule
\end{tabular}
\vspace{-1ex}
\caption{Performance comparison of Tokenizer+DiT-B on the DAIN-HD544p dataset.}
\label{tab:pffm_dit}
\end{table}

\vspace{-4ex}
\subsection{Different training datasets}
We provide the results of LDMVFI and LBBDM trained on the same dataset, LAVIB, in \cref{tab:lavib_comparison}. Clearly, EDEN still outperforms them when using the same training data.
\begin{table}[h]
\centering
\begin{tabular}{lcccc}
\toprule
 & PSNR$\uparrow$    & SSIM$\uparrow$    & LPIPS$\downarrow$   & FloLPIPS$\downarrow$ \\ \midrule
LDMVFI                  & 25.88 & 0.937  & 0.1501  & 0.2413   \\
LBBDM                   & 26.56 & 0.944  & 0.1477  & 0.2366   \\
EDEN                    & \textbf{26.85} & \textbf{0.944}  & \textbf{0.1321}  & \textbf{0.2187}   \\ \bottomrule
\end{tabular}
\vspace{-1ex}
\caption{Performance comparison with the same training dataset on the DAIN-HD544p dataset.}
\label{tab:lavib_comparison}
\end{table}

\vspace{-4ex}
\subsection{Same number of diffusion steps}
\cref{tab:denoising_runtime} shows the results on DAIN-HD544p using the same number of denoising steps. Clearly, EDEN achieves higher performance with faster speed compared to both LBBDM and LDMVFI.
\begin{table}[ht]
\centering
\resizebox{1.0\linewidth}{!}{
\begin{tabular}{lcccc}
        \toprule
          & Denoising Steps & LPIPS $\downarrow$ & FloLPIPS $\downarrow$ & RT (s) $\downarrow$ \\
        \midrule
        LDMVFI  & \multirow{3}{*}{2} & 0.1501 & 0.2413 & 0.525 \\
        LBBDM   &                    & 0.1477 & 0.2366 & 0.907 \\
        EDEN    &                    & \textbf{0.1321} & \textbf{0.2187} & \textbf{0.250} \\
        \bottomrule
    \end{tabular}}
\vspace{-1ex}
\caption{Runtime comparison of various methods (for interpolating a 544x1280 frame) at 2 denoising steps.}
\label{tab:denoising_runtime}
\end{table}

\vspace{-4ex}
\subsection{Limitations and Feature Work}
\begin{figure}[h]
    \centering
    \includegraphics[width=1.0\linewidth]{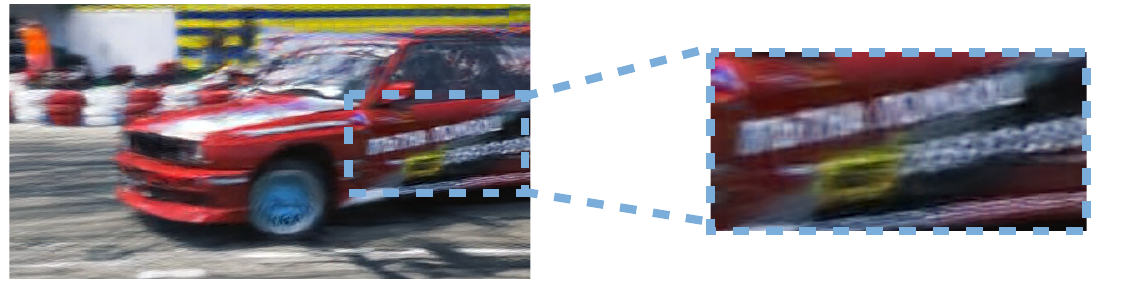}
    \vspace{-4ex}
    \caption{Visualization of results with blurred text.}
    \label{fig:limitations}
\end{figure}

Though our method demonstrates significant improvements in handling complex motions, it still has certain limitations. Specifically, it struggles with blurring when dealing with rapid changes in fine details (e.g., text). As illustrated in \cref{fig:limitations}, while our method accurately captures the positions of moving car, the text appear blurred. A possible reason for this limitation is that our decoder directly applies pixel shuffle on the tokenizer decoder final layer's output to generate the image, which inherently introduces some degree of blurring. In future work, we plan to explore an effective pixel decoder network to transform the final output of the tokenizer decoder into sharper, more realistic images.
\vspace{-4ex}

\subsection{More Visualizations}
\begin{figure*}[h]
    \centering
    \includegraphics[width=1.0\linewidth]{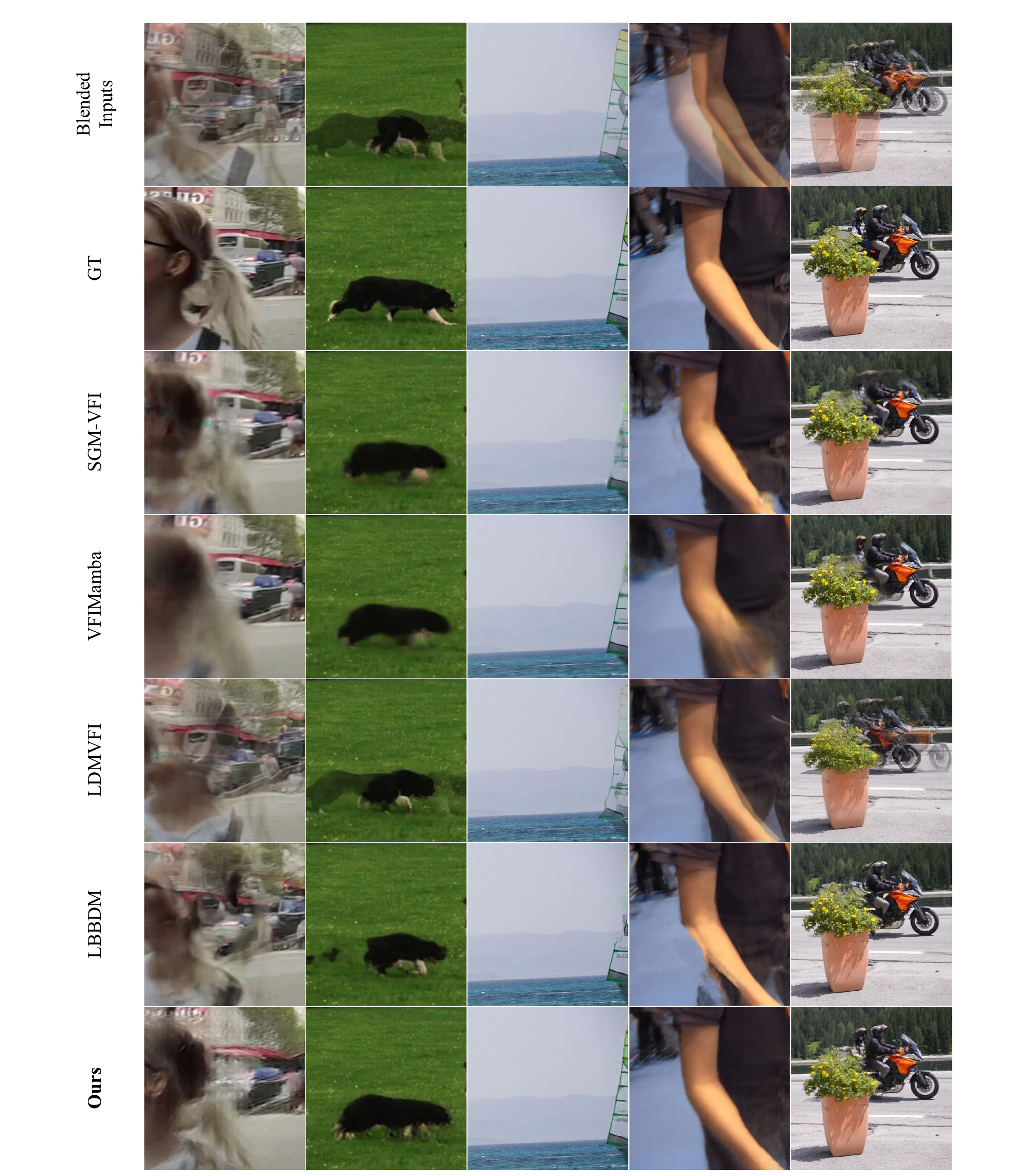}
    \vspace{-4ex}
    \caption{Visual comparison with different methods, examples selected from DAVIS. Ours outperforms previous methods in both capturing the motion of multiple objects and modeling fast, nonlinear motions.}
    \label{fig:supp_visual}
\end{figure*}

We provide additional visualization comparisons against previous state-of-the-art methods in \cref{fig:supp_visual}. These results demonstrate that our method effectively handles complex or nonlinear motions in video frame interpolation. In comparison, prior methods struggle to accurately model such motions.
